# WTCL-DEHAZE: RETHINKING REAL-WORLD IMAGE DEHAZING VIA WAVELET TRANSFORM AND CONTRASTIVE LEARNING

Divine Joseph Appiah, Donghai Guan, Abdul Nasser Kasule and Mingqiang Wei

School of Computer Science and Technology, Nanjing University of Aeronautics and Astronautics

*ABSTRACT*

*Images captured in hazy outdoor conditions often suffer from colour distortion, low contrast, and loss of detail, which impair high-level vision tasks. Single image dehazing is essential for applications such as autonomous driving and surveillance, with the aim of restoring image clarity. In this work, we propose WTCL-Dehaze an enhanced semi-supervised dehazing network that integrates Contrastive Loss and Discrete Wavelet Transform (DWT). We incorporate contrastive regularization to enhance feature representation by contrasting hazy and clear image pairs. Additionally, we utilize DWT for multi-scale feature extraction, effectively capturing high-frequency details and global structures. Our approach leverages both labelled and unlabelled data to mitigate the domain gap and improve generalization. The model is trained on a combination of synthetic and real-world datasets, ensuring robust performance across different scenarios. Extensive experiments demonstrate that our proposed algorithm achieves superior performance and improved robustness compared to state-of-the-art single image dehazing methods on both benchmark datasets and real-world images.*

*KEYWORDS*

*Image Dehazing, Semi-Supervised Learning, Contrastive Loss, Discrete Wavelet Transform, & Deep Learning*

## 1. INTRODUCTION

Images captured in hazy outdoor conditions often suffer from quality issues such as colour distortion, low contrast, and loss of details. These degradations can significantly impair the performance of subsequent high-level vision tasks. Restoring clarity to images obscured by haze is a fundamental objective in computer vision, known as single image dehazing. This restoration is essential for various applications such as autonomous driving, surveillance, and remote sensing. The formation of a hazy image can be modelled by the atmospheric scattering model:

$$I(x) = J(x)t(x) + A(1 - t(x)) \qquad (1)$$

Where $I(x)$ is the observed hazy image, $J(x)$ is the scene radiance (clear image), $t(x)$ is the transmission map, and $A$ is the global atmospheric light. The transmission map $t(x)$ describes the portion of light that reaches the camera without being scattered and is given by:

$$t(x) = e^{-\beta d(x)} \qquad (2)$$

where $\beta$ is the scattering coefficient and $d(x)$ is the scene depth. The goal of dehazing is to estimate $J(x)$ from $I(x)$, $A$, and $t(x)$. As only the hazy image $I(x)$ is available, the problem is inherently ill-posed.

To make this problem well-posed, existing methods usually make assumptions about the clean images. For instance, the Dark Channel Prior (DCP), introduced by He et al. [31], is based on the finding that the lowest intensity in a small region of most haze-free images is usually very low. This method estimates the transmission map and atmospheric light to recover the clear image. Similarly, Zhu et al. [2] developed the Colour Attenuation Prior (CAP), which estimates haze levels by examining the correlation between pixel brightness and saturation to restore clarity.

However, since image priors often involve non-convex and non-linear terms, these approaches entail high computational loads. With the advent of deep learning, many CNN-based methods have been developed for image dehazing to overcome these limitations. Ren et al. [4] introduced a multi-scale CNN for estimating transmission maps in a coarse to fine manner. Zhang and Patel [5] proposed a densely connected pyramid dehazing network that combined CNN with the U-Net architecture for improved performance. Recent advances include Qu et al. [34], who used generative adversarial networks (GANs) to directly generate clear images from hazy ones, leveraging adversarial training for better quality restoration. Chen et al. [7] developed an unsupervised dehazing method using GANs to address the domain gap between synthetic and real-world images. Semi-supervised learning has been explored to utilize both labelled and unlabelled data. Wang et al. [8] proposed a framework that incorporates both types of data to train a dehazing network, enhancing its generalization across different image domains. Zhang and Li [9] used adversarial training in a semi-supervised setting to improve model robustness. More recent works include Chen et al. [40], who fine-tuned models using real hazy images in an unsupervised manner, and Zhang et al. [11], who developed Semi-DerainGAN using both synthetic and real data for improved performance. Transformers have recently shown potential in computer vision tasks. Vision Transformers (ViTs) and their variants, such as SwinIR and Uformer, have been adapted for low-level vision tasks, including dehazing [12, 13]. These methods leverage the attention mechanism to capture global context and multi-scale information, which is crucial for effective dehazing. In this paper, we build upon the existing Semi-Supervised dehazing network by exploring Contrastive Loss and Discrete Wavelet Transform (DWT) along with Inverse Discrete Wavelet Transform (IDWT). Our key contributions are as follows:

- We introduce contrastive regularization to improve the feature representation by contrasting hazy and clear image pairs, enhancing the dehazing performance.

- We integrate wavelet transform into the network architecture, allowing for effective multi-scale feature extraction, capturing both high-frequency details and global structures.

- Our approach leverages both labeled and unlabeled data, trained on synthetic and real datasets, to bridge the domain gap and improve generalization.

We conduct extensive experiments to evaluate the performance of our proposed method. The results demonstrate that our approach outperforms several state-of-the-art dehazing methods on standard benchmarks on both synthetic and real hazy images.

## 2. RELATED WORK

### 2.1. Prior-Based Single Image Dehazing

Prior-based methods leverage physical models and statistical properties of haze to restore clear images. He et al. [31] introduced the Dark Channel Prior (DCP), based on the observation that in most non-sky regions of haze-free images, at least one color channel has very low intensity. This prior allows for effective estimation of the transmission map and atmospheric light, leading to significant improvements in dehazing performance. Zhu et al. [2] proposed the Color Attenuation Prior (CAP), which estimates haze density by exploiting the relationship between brightness and saturation of pixels. CAP has been effective in producing clear images from hazy inputs. Fattal [3] developed an approach based on the observation that small image patches exhibit a 1D distribution in the RGB color space, which helps in estimating the transmission and atmospheric light for haze removal. Ren et al. [4] enhanced traditional methods by introducing a Multi-Scale Convolutional Neural Network that integrates spatial priors into the transmission map estimation process, thereby capturing fine-grained details in hazy scenes. Prior-based methods rely heavily on assumptions and handcrafted priors. While effective, these methods often struggle with scenes that do not conform to these assumptions, leading to artifacts and reduced performance in diverse environments. The computational complexity associated with non-convex and non-linear terms also limits their efficiency and scalability.

### 2.2. Learning-Based Single Image Dehazing

Deep learning has significantly advanced the field of image dehazing. Ren et al. [4] proposed a coarse-to-fine strategy for learning the mapping from hazy inputs to transmission maps. Zhang and Patel [5] introduced a densely connected pyramid network for transmission map estimation, which used a U-Net for estimating atmospheric light, leading to improved performance. More recent work includes methods like Qu et al. [34], which employs a GAN-based approach for image dehazing, leveraging adversarial training to directly generate clear images from hazy inputs. Additionally, Chen et al. [7] proposed an unsupervised dehazing method using GANs to handle the domain gap between synthetic and real-world data, showing significant improvements in performance. Further advancements have seen the use of Vision Transformers. Vision Transformers have shown promise in replacing traditional CNNs for various tasks, including dehazing. DehazeFormer, a novel transformer-based architecture, adapts Swin Transformer for image dehazing by incorporating several enhancements to better handle haze removal [12]. SwinIR and Uformer are other examples of transformer-based networks tailored for low-level vision tasks, including dehazing [13]. Learning-based methods, particularly those leveraging deep CNNs and GANs have shown significant improvements in dehazing performance. However, these methods typically require large amounts of labeled training data, which can be challenging to obtain for real-world hazy images. The reliance on synthetic datasets can introduce domain gaps, causing these models to perform suboptimally on real-world images that differ from the training data [4,7].

### 2.3. Semi-Supervised Learning for Image Dehazing

Semi-supervised learning approaches leverage both labeled and unlabeled data, improving model performance and generalization. Wang et al. [8] proposed a framework that integrates labeled and unlabeled data for training a dehazing network, enhancing generalization across diverse image domains. Zhang and Li [9] used adversarial training in a semi-supervised context to improve the robustness of dehazing models. Chen et al. [40] introduced a synthetic-to-real dehazing network that fine-tunes models using real hazy images in an unsupervised manner to bridge the domain gap between synthetic and realworld data. Similarly, Zhang et al. [11] developed Semi-DerainGAN, a GAN-based method that utilizes both synthetic and real data for improved dehazing performance in real-world scenarios.

### 2.4. Contrastive Learning

Contrastive learning has shown great potential in improving feature representations by contrasting positive and negative pairs. It aims to pull the anchor close to positive points and push the anchor far away from negative points in the representation space. Wu et al. [14] introduced a contrastive regularization term that leverages both hazy and clean images to enhance dehazing performance, demonstrating the effectiveness of contrastive learning in low-level vision tasks. Han et al. [15] applied contrastive learning to underwater image restoration, further highlighting its potential to improve feature representations and overall restoration quality. Additionally, Chen et al. [7] proposed an unsupervised contrastive learning framework for image dehazing, which aligns the features of hazy and haze-free images using a contrastive loss, thereby pushing the boundaries of dehazing performance without relying on paired data.

### 2.5. Wavelet Transform

The use of wavelet transforms in neural networks has been explored to enhance feature extraction capabilities by capturing multi-scale information. Liu et al. [16] demonstrated the effectiveness of wavelet transforms in improving the image super-resolution by decomposing the image into different frequency components and processing them separately. Similarly, Huang et al. [17] incorporated wavelet transforms into dehazing networks to capture high-frequency details and improve the visual quality of restored images. Wavelet transforms have been particularly useful in image restoration tasks, allowing networks to process multi-scale information more effectively. This capability is crucial for tasks like dehazing, where both global structures and fine details need to be preserved and enhanced.

## 3. METHODOLOGY

Typically, deep CNN-based networks for image dehazing are trained using supervised learning methods, often limited to synthetic datasets. This reliance on synthetic data often fails to generalize well to real-world scenarios due to the domain gap. To overcome this limitation, we introduce a semi-supervised learning approach for image dehazing. Our approach leverages both labeled and unlabeled datasets to enhance the model's generalization capabilities.

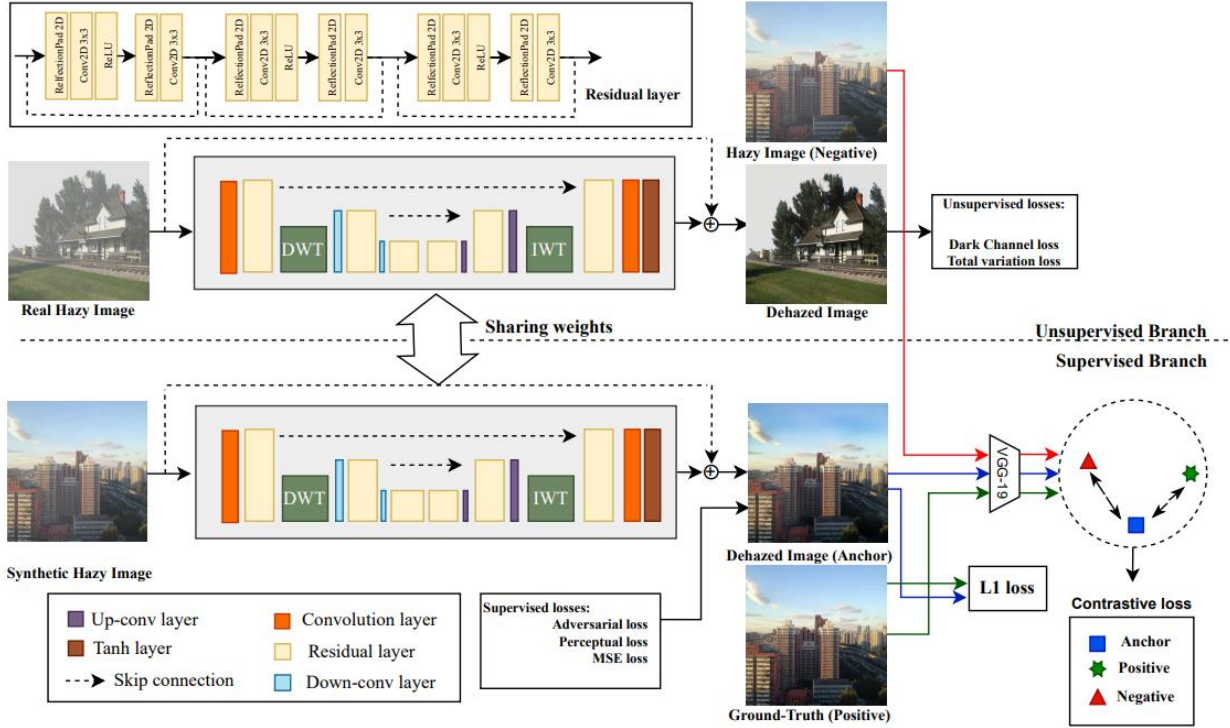

Figure 1. Overview of WTCL-Dehaze. WTCL-Dehaze is a semi-supervised learning framework designed for single-image dehazing. The method features two branches that share weights, using a residual layer with no normalization. The supervised branch is trained with labeled synthetic data, utilizing the supervised losses and Contrastive loss (which includes the L1 loss). Meanwhile the unsupervised branch is trained with unlabeled real data employing the unsupervised losses.

Specifically, we train our model for image dehazing using labeled dataset $\{I_i, J_i\}_{i=1}^{N_l}$ and unlabeled dataset $\{I_i\}_{i=1}^{N_{un}}$, where $N_l$ and $N_{un}$ denote the numbers of the labeled and unlabeled training images respectively. Here, $I_i$ and $J_i$ represent the $i$-th hazy image and its corresponding ground-truth clean image. The objective of our training process is to enable the model to generate haze-free images $J$ from hazy inputs $I$:

$$J = G(I) \qquad (3)$$

where $G(\cdot)$ represents the proposed network architecture. This architecture comprises a supervised branch $G_s$ and an unsupervised branch $G_{un}$. Both branches share the same weights during the training process to ensure consistency and to leverage the strengths of supervised and unsupervised learning. The supervised branch $G_s$ focuses on learning from the labeled data by minimizing the reconstruction loss between the generated images and the ground-truth images. Meanwhile, the unsupervised branch $G_{un}$ utilizes the unlabeled data to enforce consistency in the learned representations, encouraging the model to produce realistic and high-quality dehazed images even in the absence of ground-truth labels.

## 3.1 Network Architecture

Our proposed network builds upon an existing semi-supervised deahazing framework. The baseline architecture includes an encoder-decoder structure with skip connections, designed to handle multi-scale feature extraction. [23–25]. The architecture and configurations of our proposed network are illustrated in Figure 1. The encoder comprises three scales, each containing three residual blocks without normalization layers, as per the approach of [26]. The decoder mirrors the encoder's structure, with up-sampling achieved through Transposed-Conv layers. Every convolution layer is followed by a ReLU activation, except for the final layer (Conv24).To down-sample the feature maps from the previous scale by a factor of 1/2, we use a Stride-Conv layer. Feature maps are skip-connected through summation operations. Additionally, residual learning is utilized to learn the difference between hazy and clean images. In the innermost layer of the encoder-decoder architecture, we integrate the Discrete Wavelet Transform (DWT) for multi-scale feature extraction. The DWT decomposes the feature maps into details and global structures. After processing through several residual blocks, the Inverse Discrete Wavelet Transform (IWT) is applied to reconstruct the feature maps from the wavelet coefficients. This integration helps preserve and enhance both global structures and fine details in the dehazed images. For adversarial learning, we construct a discriminator using conventional classifier architecture, composed of convolution, non-linear ReLU, and instance normalization layers. Additionally, we incorporate contrastive loss into the network. During the training, the contrastive loss function is applied to both the supervised and unsupervised branches. It improves feature representation by contrasting hazy and clean image pairs, enhancing the dehazing performance of our network. The contrastive loss is calculated using the real hazy image, ground-truth clean image and it is added to the total generator loss.

## 3.2 Training Losses

In this section, we outline the different types of losses used during training.While we employ serveral established losses to ensure effective training, our key contribution is the introduction of contrastive regularization.

*Supervised Losses Mean Squared Loss*: We utilize the mean squared loss to ensure the proximity between the predicted image $J$ and the ground-truth image $J$ :

$$L_{MSL} = \frac{1}{N_l} \sum_{i=1}^{N_l} \left\| J_i - J_i \right\|_2 \tag{4}$$

where $N_l$ denotes the number of labeled data in a mini-batch.

*Perceptual Loss*: We use the pre-trained VGG-19 network for the perceptual loss function to generate photo-realistic images:

$$L_{PL} = \frac{1}{N_l} \sum_{i=1}^{N_l} \left\| F(J_i) - F(J_i) \right\|_2 \tag{5}$$

where $F(J_i)$ and $F(J_i)$ represent the vector forms of the feature maps according to the predicted image and its corresponding ground-truth respectively. The feature maps are from the conv3-3 layer of the VGG-19 network inspired by ImageNet.

*Adversarial Loss*: To produce images with sharpness and visual appeal, we use the GAN model and the discriminator $D$ is to differentiate images produced by the generator and images from the ground-truth of the labeled data. The adversarial loss can be expressed as:

$$L_{adv} = E_J [\log D_{is}(J)] + E_J [\log(1 - D_{is}(J))] \tag{6}$$

*Unsupervised Losses Total Variation Loss*: To preserve the structures and features of the predicted images, we apply an L1 regularization gradient prior in the unsupervised branch:

$$L_{TV} = \frac{1}{N_{un}} \sum_{i=1}^{N_{un}} (\|\nabla_h J_i\|_1 + \|\nabla_v J_i\|_1) \tag{7}$$

where $\nabla_h$ and $\nabla_v$ denote the horizontal and vertical differential operation matrices respectively.

*Dark Channel Loss*: We use the L1 regularization to constrain the sparsity of predicted images. It can be expressed as:

$$L_{DC} = \frac{1}{N_{un}} \sum_{i=1}^{N_{un}} (\|D_{J_i}\|_1) \tag{8}$$

where $D_{J_i}$ denotes the vector form of the dark channel of the predicted image $J_i$. Although the dark channel prior has been shown effective for haze removal by adding constraints on the clean images, its highly non-convex and non-linear nature makes it challenging to embed into learning networks. We address this by applying a lookup table scheme to handle the forward and backward steps of the dark channel operation. In the forward stage, we compute the dark channel $D(I)$ of a single-channel image using a $5 \times 5$ matrix, with a patch size $N(y)$ set to 3 x 3:

$$D(I) = \min_{y \in N(x)} I(y) \tag{9}$$

*Contrastive Regularization (CR)* Contrastive learning has become a powerful tool in self-supervised learning because it effectively learns representations by pulling similar (positive) samples together and pushing dissimilar (negative) samples apart in the latent space. This method has been successfully applied in various computer vision tasks, including image classification, object detection, and low-level vision tasks like image denoising and deblurring. Inspired by Wu et al. [14], we integrate Contrastive Regularization into the supervised branch of our network. This regularization contrasts hazy and clear image pairs, enhancing feature representation by ensuring that the restored image is closer to the clear image and further from the hazy image in the latent space. Positive pairs are formed by a clear image $J$ and its restored image $J^*$ and a hazy image $I$. The restored image is referred to as the anchor, the clear image as positive and the hazy image as negative. The latent feature space is derived from intermediate features from a fixed pre-trained model $G$, such as VGG-19. The CR loss function is defined:

$$\min \|J - \phi(I, w)\|_1 + \beta \cdot \rho(G(I), G(J), G(\phi(I, w))) \tag{10}$$

where the first term is the reconstruction loss aligning the restored image with its groundtruth, using the L1 loss for better performance compared to L2 loss. The second term, $\rho(G(I), G(J), G(\phi(I, w)))$, is the contrastive regularization among $I$, $J$, and $\phi(I, w)$ in the latent feature space, balancing the forces pulling the restored image $\phi(I, w)$ towards the clear image $J$ and pushing it away from the hazy image $I$. $\beta$ is a hyperparameter for balancing these losses. To ensure the contrastive ability, hidden features from different layers of the fixed pre-trained model are used. This contrastive loss is applied in both the supervised and unsupervised branches, allowing the network to leverage labeled and unlabeled data effectively:

$$\min \|J - \phi(I, w)\|_1 + \beta \sum_{i=1}^{n} w_i \cdot D(G_i(J), G_i(\phi(I, w))) \\ - D(G_i(I), G_i(\phi(I, w))) \tag{11}$$

where $G_i$ extracts the $i$-th hidden features, $D(x, y)$ is the L1 distance between $x$ and $y$, and $w_i$ is the weight coefficient. While L1 loss reduces pixel-wise errors, it may miss crucial perceptual details. By incorporating contrastive learning, our approach ensures the model captures both pixel-level accuracy and meaningful feature space relationships, leading to clearer and more realistic dehazed images. The overall loss function combines supervised, unsupervised losses, adversarial, and contrastive losses to train the proposed network. It is formulated as follows:

$$L = L_{MSL} + \alpha L_{PL} + \beta L_{TV} + \gamma L_{DC} + \delta L_{adv} + \epsilon L_{cont} \qquad (12)$$

where $\alpha$, $\beta$, $\gamma$, $\delta$, $\epsilon$ are the positive weights of each loss component. These hyperparameters balance the contributions of different losses to the overall objective.

### 3.3 Discrete Wavelet Transform

The ability of DWT to separate high-frequency details, such as edges and textures, from low-frequency structures, such as smooth regions, is particularly advantageous in dehazing tasks. The 2D Discrete Wavelet Transform (DWT) serves as a potent mechanism for multiscale feature extraction in image processing. Using the Haar wavelet as a case study, the 2D DWT decomposes an image into distinct frequency components, capturing both highfrequency and low-frequency structures [18–20]. This method applies a sequence of highpass and low-pass filters to the image. The 2D DWT divides the image into four sub-bands i.e. Low-pass filter ($f_{LL}$) and three high-pass filters ($f_{LH}, f_{HL}, f_{HH}$). These filters have fixed parameters and employ stride-2 convolution operations during the transformation. By convolving with each filter, images are decomposed into four sub-bands: $x_{LL}$, $x_{LH}$, $x_{HL}$, and $x_{HH}$. The sub-band $x_{LL}$ can be expressed as $(f_{LL} * x) \downarrow 2$, where * denotes the convolution operation, $x$ is the input signal, and $\downarrow 2$ indicates down-sampling by a factor of 2. Using Haar wavelet as an example, the four filters can be defined as:

$$\begin{aligned} f_{LL} &= \begin{pmatrix} 1 & 1 \\ 1 & 1 \end{pmatrix}, \\ f_{LH} &= \begin{pmatrix} -1 & -1 \\ 1 & 1 \end{pmatrix}, \\ f_{HL} &= \begin{pmatrix} -1 & 1 \\ -1 & 1 \end{pmatrix}, \\ f_{HH} &= \begin{pmatrix} 1 & -1 \\ -1 & 1 \end{pmatrix}, \end{aligned} \qquad (13)$$

The ($i$, $j$)-th value of $x_{LL}$ after the 2D Haar wavelet transform is defined as:

By applying DWT, we decompose the image into frequency components, which allows the model to learn distinct features from both high-frequency and low-frequency information. This single application of DWT enhances the model's ability to differentiate between detailed textures and broader structures, ultimately improving the overall dehazing performance. However, applying DWT alone isn't sufficient to achieve optimal results. Therefore, we combine the frequency-domain operations with convolutional layers, which process the sub-bands separately. This approach allows the network to learn both spatial and frequency information, improving its ability to capture and restore image details. [21,22]. The Inverse Wavelet Transform (IWT) is then used to reconstruct the original image from the four sub-bands $[LL, LH, HL, HH] \to I$. This reconstruction step is crucial for combining the detailed information learned from each sub-band, resulting in a high-quality, dehazed image that retains both fine details and overall structure, this can be formulated as:

$$\begin{aligned} x_{LL}(i, j) &= x(2_i - 1, 2_j - 1) + x(2_i - 2, 2_j) \\ &+ x(2_i, 2_j - 1) + x(2_i, 2_j) \end{aligned} \qquad (14)$$

Following the methodology of [27], we incorporate the Discrete Wavelet Transform into the encoder to decompose feature maps into multiple scales, capturing both high-frequency details and global structures. The Inverse DWT (IWT) is applied in the decoder to reconstruct the feature maps. The decomposition allows the network to isolate and process fine details separately from larger structural elements. The Inverse Wavelet Transform (IWT) is applied within the decoder to reconstruct the feature maps from the wavelet coefficients.

## 4. EXPERIMENTS

### 4.1 Implementation Details

We alternate updates between the generator and discriminator, updating the discriminator once after every five generator updates. The generator is optimized in a semi-supervised fashion. We utilize the PyTorch toolbox [28] and the Adam [29] solver for both components with parameters set to $\beta_1 = 0.9$, $\beta_2 = 0.99$, and a weight decay of of $10^{-4}$. Training occurs over 300 epochs, starting with a learning rate of $10^{-4}$ for the first 150 epochs, then decreasing linearly to $10^{-6}$ over the next 150 epochs using the formula:

$$lr = 10^{-4} - \frac{10^{-4} - 10^{-6}}{150} \times (E - 150) \quad (15)$$

where $E$ denotes the number of the training epoch. We crop the images to the size of $256 \times 256$ and normalize the pixel values to [-1,1]. We set the patch size as $256 \times 256$ when computing DC loss. The loss weights are set as: $\alpha = 10^{-2}$, $\beta = 10^{-5}$, $\gamma = 10^{-5}$, $\delta = 10^{-3}$, and $\epsilon = 10^{-1}$. We train our network on Nvidia GeForce RTX 3090 Ti and it takes three days to converge.

*Datasets* We randomly choose labeled and unlabeled images from the RESIDE dataset[30]. RESIDE is a widely used dataset for image dehazing, which contains six subsets, i.e ITS (Indoor Training Set), SOTS (Synthetic Object Testing), HSTS (Hybrid Subjective Testing Set), RTTS (Real-world Task-driven Testing Set), OTS (Outdoor dehazing), and URHI (Unannotated Real Hazy Images). In our experiment, we used 4000 labeled images, 2000 from the ITS dataset, and 2000 from the OTS dataset. For the unlabeled images, we randomly select 2000 real hazy images from the URHI dataset.

*Table 1.* Quantitative Results for WTCL-Dehaze and 11 leading Dehazing methods on Synthetic Datasets

| Method | Publication | Type | SOTS outdoor | | HSTS | |
|---|---|---|---|---|---|---|
| | | | PSNR | SSIM | PSNR | SSIM |
| DCP | TPAMI'11 | Prior | 18.38 | 0.819 | 17.01 | 0.803 |
| NCP | CVPR'16 | Prior | 18.07 | 0.802 | 17.62 | 0.798 |
| AOD-Net | ICCV'17 | Supervised | 20.08 | 0.861 | 19.68 | 0.835 |
| EPDN | CVPR'19 | Supervised | 22.57 | 0.863 | 20.37 | 0.877 |
| FD-GAN | AAAI'20 | Supervised | 23.76 | 0.926 | 23.28 | 0.914 |
| Interleaved CSF | TIP'20 | Supervised | 24.17 | 0.923 | 22.94 | 0.907 |
| Semi-Dehazing | TIP'20 | Semi-Supervised | 24.79 | 0.892 | 24.36 | 0.889 |
| CycleGAN | ICCV'17 | Unsupervised | 16.05 | 0.706 | 16.05 | 0.703 |
| Cycle-Dehaze | CVPRW'18 | Unsupervised | 17.96 | 0.797 | 17.96 | 0.777 |
| YOLY | IJCV'21 | Unsupervised | 21.02 | 0.889 | 21.02 | 0.905 |
| PSD | CVPR'21 | Unsupervised | 19.37 | 0.844 | 19.37 | 0.824 |
| WTCL-Dehaze | Ours | Semi-Supervised | **27.24** | **0.971** | **27.24** | **0.918** |

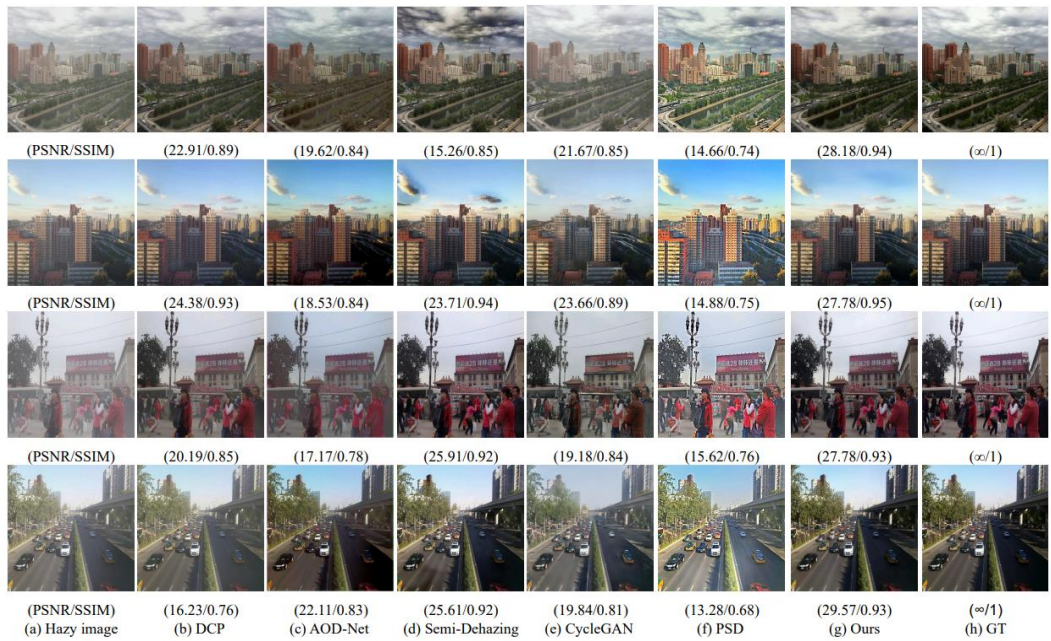

Figure 2. Image dehazing results on the SOTS outdoor dataset. From (a) to (h): (a) the hazy image, and the dehazing results of (b) DCP [8], (c) AOD-Net [33], (d) Semi-Dehazing [41], (e) CycleGAN [37], (f) PSD [40], (g) Our WTCL-Dehaze and (h) ground-truth image respectively. Our method generates cleaner results with fewer artifacts and minimal color distortion.

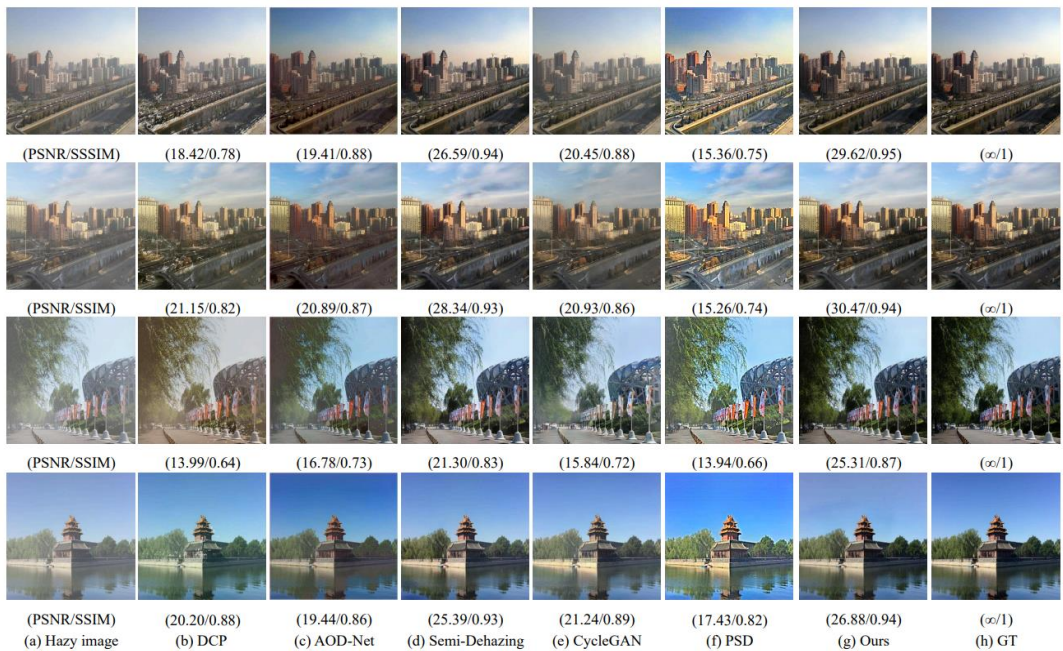

Figure 3. Image dehazing results on the HSTS outdoor dataset. From (a) to (h): (a) the hazy image, and the dehazing results of (b) DCP [8], (c)AOD-Net [33], (d)Semi-Dehazing [41], (e)CycleGAN [37], (f)PSD [40], (g) our WTCL-Dehaze and (h) ground-truth image respectively. WTCL-Dehaze can generate much clearer and visually appealing results

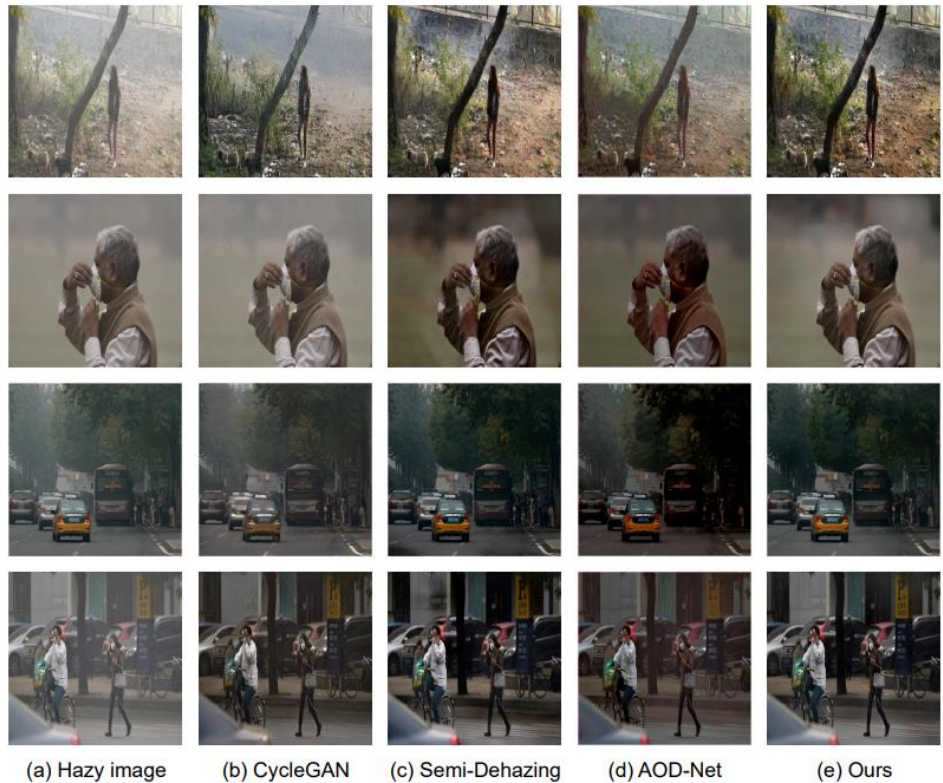

Figure 4. Image dehazing results on the Real-World outdoor dataset. From (a) to (e): (a) the hazy image, and the dehazing results of (b) CycleGAN [37], (c) Semi-Dehazing [41], (d) AOD-Net [33], (e) Our WTCL-Dehaze and (h) ground-truth image respectively. WTCL-Dehaze can produce both haze-free and more natural images.

### 4.2 Evaluation Settings

WTCL-Dehaze is compared quantitatively and qualitatively with different image dehazing approaches. They can be grouped into three categories: 1) prior-based DCP [31] and NCP [32]; 2) supervised-based AOD-Net [33], EPDN [34], FD-GAN [35], and Interleaved CSF [36]; 3) unsupervised-based CycleGAN [37], Cycle-Dehaze [38], YOLY [39], and PSD [40]. We also compared WTCL-Dehaze with Semi-dehazing [41], another semi-supervised image dehazing framework. We utilize the average Peak Signal to Noise Ratio (PSNR) and Structural Similarity Index (SSIM) for the quantitative assessment of the restored images, as these metrics are the most commonly used full-reference evaluation standards in image processing.

### 4.3 Comparison with SOTA Methods

*Quantitative comparisons* We compare our method to 11 state-of-the-art dehazing methods on SOTS outdoor and HSTS datasets in *Table 1*. Under normal circumstances haze usually affects outdoor vision systems, hence we mainly focus on evaluating methods on outdoor datasets only. For comparison and analysis, we adopted a consistent evaluation protocol where we either retrained the state-of-the-art models on the OTS dataset or used the pre-trained models provided by the authors. To ensure fairness, all models were tested using the same hazy image inputs, and we computed the Peak Signal to Noise Ratio (PSNR) and Structural Similarity Index (SSIM) to quantify image quality. Our evaluation also includes detailed visual comparisons on outdoor datasets to showcase qualitative differences between methods. Our WCL-Dehaze outperforms both supervised and semi-supervised image dehazing methods in terms of SSIM and PSNR metrics. For example, compared with Semi-Dehazing on the SOTS outdoor dataset, our method increases PSNR by *1.97* and SSIM by *0.079*.

*Visual comparisons* Figure 4 provides visual comparisons on real-world outdoor datasets, highlighting the superior effectiveness of our WTCL-Dehaze method over other state-of-the-art approaches. These examples illustrate how our method generates clearer and more natural images with fewer artifacts and minimal colour distortion, even in real-world scenarios. We observe that other methods are not sensitive enough to capture important details in the image after dehazing, resulting in a slightly hazy appearance and color distortion. We observe that our proposed WTCL-Dehaze can accurately dehaze images as closely matching the ground-truth image. Comparisons on Row 2 and Row 3 shows that our method is more effective in dehazing while preserving image details such as color, brightness and sharpness. Overall, these results confirm that our method provides the most accurate and visually appealing dehazing performance among the techniques evaluated.

## 4.4 Ablation Study

***Effect on different components in WTCL-Dehaze.*** The proposed WTCL-Dehaze network demonstrates superior dehazing performance compared to SOTA methods. To further evaluate the effectiveness of WTCL-Dehaze, we conduct ablation studies, analyzing the impact of different component on overall performance. The variants considered are as follows:

A. *Baseline + L (Baseline\*)*: This configuration represents our baseline model trained with a standard loss function $L$, which includes supervised and unsupervised losses but excludes additional enhancements such as DWT, IWT, and contrastive loss. This serves as the foundation for comparing the incremental improvements brought by each additional component.

B. *Baseline\* + DWT and IWT*: In this variant, we incorporate the Discrete Wavelet Transform (DWT) and Inverse Wavelet Transform (IWT) into the baseline model. The inclusion of DWT and IWT allows the model to capture multi-scale frequency information, which is crucial for restoring fine details and global structures in hazy images. The improvement in PSNR and SSIM over the baseline (**26.28** and **0.927**, respectively) confirms the effectiveness of integrating frequency-domain operations in enhancing the dehazing capability.

C. *Baseline\* + Contrastive Loss*: This configuration builds upon the baseline model by adding contrastive loss, which is designed to improve the feature representation by contrasting hazy and clear image pairs. The contrastive loss ensures that the latent features of dehazed images are closer to the features of clear images while being distinct from hazy ones. This addition results in a noticeable improvement in both PSNR (**26.56**) and SSIM (**0.929**), indicating that contrastive learning effectively enhances the perceptual quality of the dehazed images.

D. *Ours (Baseline\* + DWT and IWT + Contrastive Loss)*: The final configuration combines all the components—baseline, DWT and IWT, and contrastive loss. This comprehensive model leverages the strengths of both spatial and frequency-domain learning, along with enhanced feature representation through contrastive learning. The resulting PSNR (**26.76**) and SSIM (**0.971**) are the highest among all variants, demonstrating that the full WTCL-Dehaze model achieves the best dehazing performance by effectively integrating these components. All these varaints are retrained following the same procedure as before and tested on the SOTS dataset. The performance results of these variants are summarized in *Table 2* and Fig 4.

*Table 2*. Ablation Analysis on WTCL-Dehaze PSNR and SSIM results on theSOTS Outdoor Dataset. The Comparisson includes Baseline methods with different Configurations anf the Proposed method

| Method | PSNR | SSIM |
|---|---|---|
| Baseline\* + L | 26.02 | 0.925 |
| Baseline\* + DWT & IWT | 26.28 | 0.927 |
| Baseline\* + Contrastive loss | 26.56 | 0.929 |
| Ours | **26.76** | **0.971** |

International Journal on Cybernetics & Informatics (IJCI) Vol.13, No.5, October 2024## 5. CONCLUSION

In this paper, we presented WTCL-Dehaze, a novel semi-supervised network for singleimage dehazing that leverages the strengths of contrastive learning and discrete wavelet transforms. Through extensive experiments, we demonstrated that WTCL-Dehaze achieves superior performance and enhanced robustness, outperforming several state-of-the-art methods on both synthetic and real-world datasets in terms of PSNR and SSIM. Our ablation studies confirmed the effectiveness of each component of our model, highlighting the importance of integrating both multi-scale feature extraction and contrastive regularization for improved dehazing performance.

Moving forward, we plan to explore further enhancements to the model, such as integrating more advanced transformers or exploring other forms of multi-scale analysis. Additionally, applying our approach to other related tasks, such as low-light image enhancement or underwater image restoration, could be valuable directions for future research.

## REFERENCES

[1] K. He, J. Sun, and X. Tang. "Single image haze removal using dark channel prior." IEEE Transactions on Pattern Analysis and Machine Intelligence, 33(12), 2341-2353, 2011.

[2] Q. Zhu, J. Mai, and L. Shao. "A fast single image haze removal algorithm using color attenuation prior." IEEE Transactions on Image Processing, 24(11), 3522-3533, 2015.

[3] R. Fattal. "Single image dehazing." ACM Transactions on Graphics (TOG), 27(3), 1-9, 2008.

[4] W. Ren, S. Liu, H. Zhang, J. Pan, X. Cao, and M. H. Yang. "Single image dehazing via multi-scale convolutional neural networks." In European Conference on Computer Vision, pp. 154- 169, 2016.

[5] H. Zhang and V. M. Patel. "Densely connected pyramid dehazing network." In Proceedings of the IEEE Conference on Computer Vision and Pattern Recognition, pp. 3194-3203, 2018.

[6] Y. Qu, Y. Chen, J. Huang, and Y. Xie. "Enhanced Pix2pix dehazing network." In Proceedings of the IEEE/CVF Conference on Computer Vision and Pattern Recognition, pp. 8160-8168, 2019.

[7] Z. Chen, Y. Wang, Y. Yang, D. Liu, and M. Cheng. "Unsupervised domain adaptation for dehazing." In Proceedings of the IEEE/CVF Conference on Computer Vision and Pattern Recognition, pp. 3106-3115, 2020.

[8] Y. Wang, H. Wang, S. Wang, M. Li, and Y. Qiao. "A novel semi-supervised deep learning framework for single image dehazing." In Proceedings of the IEEE International Conference on Computer Vision, pp. 3254-3262, 2017.

[9] J. Zhang and S. Li. "Semi-supervised single image dehazing using adversarial learning." In Proceedings of the IEEE International Conference on Image Processing, pp. 3325-3329, 2019.

[10] Z. Chen, Y. Wang, Y. Yang, D. Liu, and M. Cheng. "PSD: Principled synthetic-to-real dehazing guided by physical priors." In Proceedings of the IEEE/CVF Conference on Computer Vision and Pattern Recognition, pp. 7180-7189, 2021.

[11] H. Zhang, V. M. Patel, and R. Mohan. "Semi-DerainGAN: Semi-supervised deraining using generative adversarial networks." In Proceedings of the IEEE/CVF Conference on Computer Vision and Pattern Recognition, pp. 2382-2391.

[12] J. Li, P. Liu, W. Chen, and S. Zheng. "Dehaze-Former: Transformer-based image dehazing." arXiv preprint arXiv:2204.03883, 2022.

[13] X. Li, X. Wu, H. Lin, and W. Liu. "Uformer: A U-shaped transformer for image restoration." In Proceedings of the IEEE/CVF Conference on Computer Vision and Pattern Recognition, pp. 1763-1773, 2021.

[14] Y. Wu, X. Liu, C. Shen, and J. Yang. "Contrastive learning for low-level vision." In Proceedings of the IEEE/CVF Conference on Computer Vision and Pattern Recognition, pp. 10136-10145, 2021.

[15] J. Han, P. Li, J. Dong, Z. Li, and H. Tang. "Underwater image restoration via contrastive learning." IEEE Transactions on Image Processing, 30, 3614-3626, 2021.

[16] D. Liu, Y. Wang, X. Fan, and Z. Liu. "Image super-resolution using convolutional neural networks." In Proceedings of the IEEE/CVF Conference on Computer Vision and Pattern Recognition, pp. 2780-2788, 2018.96

International Journal on Cybernetics & Informatics (IJCI) Vol.13, No.5, October 2024[17]    X. Huang, Y. Liang, J. Zhang, and Y. Li. "Wavelet-based image dehazing using adaptive enhancement and gamma correction." IEEE Access, 7, 89365-89375, 2019.

[18]    I. Daubechies. "Ten lectures on wavelets." SIAM, 1992.

[19]    S. Mallat. "A theory for multiresolution signal decomposition: the wavelet representation." IEEE Transactions on Pattern Analysis and Machine Intelligence, 11(7), 674-693, 1989.

[20]    A. Haar. "Zur Theorie der orthogonalen Funktionensysteme." Mathematische Annalen, 69(3), 331-371, 1910.

[21]    G. Strang and T. Nguyen. "Wavelets and filter banks." SIAM, 1996.

[22]    C. K. Chui. "An introduction to wavelets." Academic Press, 1992.

[23]    W.-S. Lai, J.-B. Huang, N. Ahuja, and M.-H. Yang, Deep Laplacian Pyramid Networks for Fast and Accurate Super-Resolution, IEEE Conference on Computer Vision and Pattern Recognition (CVPR), 2017, pp. 5835-5843. https://arxiv.org/ abs/1704.03915.

[24]    Z. Shen, Y. Zeng, Z. Fang, and M. Zhang, Deep Learning for Single Image Super-Resolution: A Brief Review, IEEE Transactions on Multimedia, vol. 20, no. 10, 2018, pp. 2454-2466.

[25]    X. Tao, H. Gao, X. Shen, J. Wang, and J. Jia, Scale-Recurrent Network for Deep Image Deblurring, IEEE Conference on Computer Vision and Pattern Recognition (CVPR), 2018, pp. 8174- 8182.

[26]    S. Nah, T. H. Kim, and K. M. Lee, Deep Multi-Scale Convolutional Neural Network for Dynamic Scene Deblurring, IEEE Conference on Computer Vision and Pattern Recognition (CVPR), 2017, pp. 257-265. https://arxiv.org/abs/1612.02177

[27]    Zhang, Y., Tian, Y., Kong, Y., Zhong, B., Fu, Y. (2019). Low-light image enhancement via hybrid diffusion model. In Proceedings of the 28th International Joint Conference on Artificial Intelligence (IJCAI-19), pp. 3574-3580.

[28]    A. Paszke et al., "Automatic differentiation in PyTorch," in Proc. Neural Inf. Process. Syst. Workshops, Nov. 2017, pp. 1–4.

[29]    D. P. Kingma and J. Ba, "Adam: A method for stochastic optimization," Dec. 2014, arXiv:1412.6980. [Online]. Available: https://arxiv.org/abs/1412.6980

[30]    B. Li et al., "Benchmarking single-image dehazing and be   yond," IEEE Trans. Image Process., vol. 28, no. 1, pp. 492–505, Jan. 2019.

[31]    K. He, J. Sun, and X. Tang, "Single image haze removal using dark channel prior," IEEE Transactions on Pattern Analysis and Machine Intelligence, vol. 33, no. 12, pp. 2341–2353, Dec. 2011.

[32]    D. Berman, S. Avidan, et al., "Non-local image dehazing," in Proceedings of the IEEE Conference on Computer Vision and Pattern Recognition, 2016, pp. 1674–1682.

[33]    B. Li, X. Peng, Z. Wang, J. Xu, and D. Feng, "AOD-Net: All-in-one dehazing network," in Proceedings of the IEEE International Conference on Computer Vision, 2017, pp. 4770–4778.

[34]    Y. Qu, Y. Chen, J. Huang, and Y. Xie, "Enhanced pix2pix dehazing network," in Proceedings of the IEEE Conference on Computer Vision and Pattern Recognition, 2019, pp. 8160–8168.

[35]    Y. Dong, Y. Liu, H. Zhang, S. Chen, and Y. Qiao, "FD-GAN: Generative adversarial networks with fusion-discriminator for single image dehazing," in Proceedings of the AAAI Conference on Artificial Intelligence, vol. 34, 2020, pp. 10729–10736.

[36]    Q. Wu, W. Ren, and X. Cao, "Learning interleaved cascade of shrinkage fields for joint image dehazing and denoising," IEEE Transactions on Image Processing, vol. 29, pp. 1788–1801, 2020.

[37]    J. Y. Zhu, T. Park, P. Isola, and A. A. Efros, "Unpaired image-to-image translation using cycle-consistent adversarial networks," in Proceedings of the IEEE International Conference on Computer Vision, 2017, pp. 2223–2232.

[38]    D. Engin, A. Genc, and H. Kemal Ekenel, "Cycle-dehaze: Enhanced CycleGAN for single image dehazing," in Proceedings of the IEEE Conference on Computer Vision and Pattern Recognition Workshops, 2018, pp. 825–833.

[39]    B. Li, Y. Gou, S. Gu, J. Z. Liu, J. T. Zhou, and X. Peng, "You only look yourself: Unsupervised and untrained single image dehazing neural network," International Journal of Computer Vision, vol. 129, no. 5, pp. 1754–1767, 2021.
97

**Authors**

**Divine Joseph Appiah** is currently a master candidate with the College of Computer Science and Technology, Nanjing University of Aeronautics and Astronautics (NUAA), Nanjing, China. He received his B.S. degree in Aircraft Design and Engineering in 2021 from NUAA. His research interests include deep learning, image processing, and computer vision.

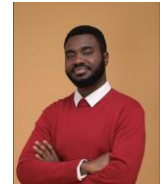

**Donghai Guan** is an associate professor with the College of Computer Science and Technology, Nanjing University of Aeronautics, Nanjing, China. He received the Ph.D. degree in Computer Engineering from Kyung Hee University (KHU), Suwon, Korea in 2009. He was a research professor in KHU during 2009-2011, assistant professor in KHU during 2012-2014. His research interests include machine learning, intelligent systems, and ubiquitous computing.

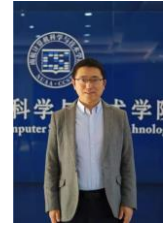

**Abdul Nasser Kasule** is currently a master candidate with the College of Civil Aviation, Nanjing University of Aeronautics and Astronautics, Nanjing, China. He received his B.S. degree in Aircraft Design and Engineering in 2021 from Nanjing University of Aeronautics and Astronautics. His research interests include computer vision, object detection, deep learning, and image processing.

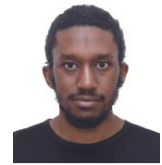

**Mingqiang Wei (Senior Member, IEEE)** received his Ph.D. degree (2014) in Computer Science and Engineering from the Chinese University of Hong Kong (CUHK). He is a professor at the School of Computer Science and Technology, Nanjing University of Aeronautics and Astronautics (NUAA). He is now an Associate Editor for ACM TOMM, The Visual Computer, and a Guest Editor for IEEE Transactions on Multimedia. His research interests focus on 3D vision, and computer graphics.

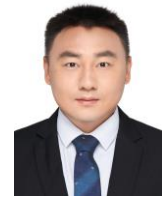